# Transfer Learning for Low-Resource Sentiment Analysis

RAZHAN HAMEED, Queen Mary University of London, England
SINA AHMADI, Insight Centre for Data Analytics, National University of Ireland Galway, Ireland
FATEMEH DANESHFAR*, University of Kurdistan, Iran

Sentiment analysis is the process of identifying and extracting subjective information from text. Despite the advances to employ cross-lingual approaches in an automatic way, the implementation and evaluation of sentiment analysis systems require language-specific data to consider various sociocultural and linguistic peculiarities. In this paper, the collection and annotation of a dataset are described for sentiment analysis of Central Kurdish. We explore a few classical machine learning and neural network-based techniques for this task. Additionally, we employ an approach in transfer learning to leverage pretrained models for data augmentation. We demonstrate that data augmentation achieves a high $F_1$ score and accuracy despite the difficulty of the task.

## 1 INTRODUCTION

Sentiment analysis has been a growing field of research that aims to classify the sentiment expressed in text automatically. The ability to accurately identify the sentiment in a text can have many applications in practice as in market research, political analysis, and social media monitoring. This problem has been widely studied in natural language processing (NLP) for some languages on various sources of data such as Twitter [2], YouTube [9, 53] and microblogs [7] across several domains as in finance [37], politics [46] and disaster management [13]. However, sentiment analysis along with other related topics, such as emotion recognition and hate speech detection, are yet to be addressed for many less-resourced languages for which collecting data and creating an evaluation dataset are challenging tasks [52].

In this paper, we focus on sentiment analysis for Central Kurdish, a less-resourced Indo-European language spoken by over 30 million speakers. One of the major limitations in the previous studies

---

*Corresponding author

Authors' addresses: Razhan Hameed, razhan@pm.me, Queen Mary University of London, 210 High Street Stratford, London, England, E15 2ZL; Sina Ahmadi, ahmadi.sina@outlook.com, Insight Centre for Data Analytics, National University of Ireland Galway, Galway Business Park, Galway, Ireland, H91 AEX4; Fatemeh Daneshfar, f.daneshfar@uok.ac.ir, University of Kurdistan, Pasdaran Boulevard, Sanandaj, Kurdistan, Iran.

on Kurdish sentiment analysis is the lack of evaluation datasets and models, making it impossible to carry out a comparative study. As such, our primary focus is to benchmark this task by manually annotating a dataset for evaluation purposes. The annotated dataset is imbalanced with remarkably more negative than positive and neutral instances. This imbalance can be problematic for machine learning models, as they may struggle to accurately classify the minority class. To remedy this issue, we also create another dataset with a more balanced distribution of sentiment using transfer learning. By the latter dataset, we are able to generate a larger dataset with a more even distribution of sentiment. Following this, we train several classical machine learning models on both datasets as well as a bidirectional long short-term memory model. Moreover, we explore the effect of emojis on sentiment analysis by training and testing our models in two different settings, with and without emojis in the text.

Our experimental results demonstrate the effectiveness of our proposed approaches and provide insights into the challenges and opportunities of sentiment analysis for Central Kurdish. Furthermore, this benchmark can pave the way for more comparative studies in the future as models and datasets are openly available at https://github.com/Hrazhan/sentiment.

## 2 RELATED WORK

With the increasing accessibility and popularity of opinion-rich resources such as online review sites, personal blogs, and social networks, opportunities, and challenges have arisen in this area. People can now use information technology to discover the opinions of others. Therefore, due to the increasing interest in systems that can directly explore thoughts and opinions, a lot of attention was drawn to the topics of opinion mining and sentiment analysis. These two fields deal with the computational encounter with the ideas, feelings, and mentality in the text. Sentiment analysis systems help gather information and provide insightful remarks from structured and unstructured online text such as email, blog posts, support tickets, web chats, social media networks, forums, and online comments.

In this section, the related work in sentiment analysis is examined. Also, previous research in this field is reviewed in general, and for the Kurdish language in particular. In addition, some of the previous methods that are useful in this field in dealing with low-resource languages are pointed out.

### 2.1 Sentiment Analysis

Opinion mining algorithms replace feeling recognition and data processing with rule-based, automated, or hybrid methods. Rule-based systems perform sentiment analysis based on predefined rules and meaningful lexicons, while automated systems learn to extract opinions using machine learning and labeled datasets. A wide range of techniques for the application of sentiment analysis has been proposed, including the hidden Markov model, Gaussian mixture model, support vector machine, neural network, contextual embeddings and zero-shot learning. In fact, there is no agreement regarding the most suitable classification method for the application of sentiment analysis and opinion mining yet.

Birjali et al. [16] review the existing methods for sentiment analysis using NLP. Among these methods, the use of techniques based on machine learning as well as statistical solutions has been beneficial. Furthermore, Rice and Zorn [48] present the resources and corpora that have been produced so far for speech sentiment analysis and shed light on how to annotate them. In this paper, the different methods that are available for evaluating the results related to the annotations of different users are compared with the same one, and the application of each one in languages with different specifications is presented, too.

With the progress in neural networks, many studies have focused on applying neural networks for the task of sentiment analysis. A new method for text emotion recognition is proposed by Basiri et al. [12] using a combination of four deep learning methods and a supervised machine learning model to recognize text emotions in annotated tweets. This model has obtained good results on the corpora of high-resource languages. Similarly, Jing et al. [27] present a framework with a new and robust model for detecting sentiment in text and stock price prediction using deep learning.

More recently, deep neural networks based on transformer architecture and contextual embeddings, such as BERT [21] and other variants such as multilingual mBART [33], have achieved state-of-the-art performance in many natural language classification tasks, including standard emotion and offensive speech recognition. Miok et al. [36] propose a Bayesian method using Monte Carlo in the attention layers of transformer models to be used to estimate the probability of hateful sentences with high reliability. Plaza-del Arco et al. [44] propose several methods for detecting text emotions in Spanish and compare their results. In this paper, the performance of models based on deep learning and machine learning, pretrained language models, the transformer architecture with attention mechanism, as well as multilingual and monolingual models of the pretrained Spanish models are compared. The performance of each one has been evaluated in terms of error analysis to understand the difficulty of the task.

Since most of the automatic sentiment analysis approaches classify the problem as binary (positive or negative), without addressing the local focus or the goal-oriented nature of the speech, Chiril et al. [19] use a multi-objective method to detect both the speech emotion and hate speech. In the presented method, a dataset labeled in one language is used to transfer knowledge to different data sets in other languages. In this model, by extracting common linguistic features in a labeled data set, it has been used to transfer this knowledge to detect offensive words in other linguistic data. In addition, topics like racism, xenophobia, sexism, misogyny, and targets of offensive words have also been identified. Also, the effect of emotional knowledge encoded in emotional computing resources (SenticNet, EmoSenticNet) and hate words with semantic structure (HurtLex) has been studied in determining specific manifestations of offensive speech.

In the case of the availability of pretrained language models with annotated sentiment analysis datasets, it is also possible to implement a zero-shot or few-shot approach for this task. For instance, Pamungkas et al. [42] use a joint learning architecture based on the zero-shot approach for cross-lingual sentiment analysis. At first, this method has been used on a rich-resource language and based on that, it has provided models for less-resource languages.

## 2.2 Kurdish Sentiment Analysis

Despite the many efforts that have been made in recent years to develop and process the Kurdish language, this language is still in the preliminary stages of development, so it has a considerable distance from the current language processing technologies[5]. Although much research has been published on tasks such as machine translation [6], tokenization [4], and developing toolkits [3], only a handful of studies address sentiment analysis for Kurdish and its varieties.[1]

As one of the earliest studies, Abdulla and Hama [1] develop a sentiment analyzer using a naive Bayes classifier with bag-of-words containing frequent words in 15,000 text documents among which 8000 are labeled as positive reviews and the rest as negative. The documents are collected from different social networks like Facebook, Twitter and Google+ and an $F_1$ measure of 0.72 is reported for this classifier. Similarly, Amin et al. [8] investigate the challenges of applying sentiment analysis approaches in the Kurdish language. These challenges are related to all stages of sentiment analysis processing from data collection to feature extraction and classification. Also, two different proposed methods, a machine learning-based method, and a lexicon-based method are presented to face these obstacles.

---

[1]An updated list of the existing tools and works on Kurdish language processing is provided at https://github.com/sinaahmadi/awesome-kurdish.

Furthermore, Awlla and Veisi [10] address sentiment analysis for Kurdish and describe the creation of a dataset containing 14,881 comments from various Facebook pages. To create an analyzer, Word2vec embeddings along with a recurrent neural network classifier are used with a reported accuracy of 71.35%. In the same vein, Azad et al. [11] address fake news detection for the Kurdish language. In this regard, a corpus is collected from a few news websites and then annotated and evaluated for the task using a few classical machine learning algorithms.

| Study | Data source | # instances | Techniques | IAA | Open-source |
|---|---|---|---|---|---|
| Abdulla and Hama [1] | Facebook, Twitter, YouTube | 15,000 | Classical ML | - | ✗ |
| Awlla and Veisi [10] | Facebook | 14,881 | Neural networks | - | ✗ |
| This work | Twitter | 1,185 (G) 4,500 (S) | classical ML, neural networks, transfer learning | 0.84 | ✓ |

Table 1. Previous works in sentiment analysis for Central Kurdish in comparison with the current paper. Inter-annotator agreement (IAA) is provided if reported in the papers. # refers to the number. Unlike the current work, none of the previous works is openly available. G and S refer to gold-standard and silver-standard data, respectively.

Table 1 summarizes the previous studies along with their datasets and techniques proposed for Kurdish sentiment analysis. Although the task has been tackled a few times, they are of little or no avail, chiefly due to the lack of open-source tools and resources. This challenge not only impedes progress in the field but also makes experimental comparisons impossible. Therefore, our main focus in this work is on the centrality of producing a benchmark for analyzing sentiments in Kurdish. For this purpose, it is necessary to produce an annotated corpus of users' opinions, by employing expert people and relying on the knowledge of those interested in this field.

## 2.3 Sentiment Analysis in Low-Resourced Scenarios

Most of the methods presented so far have been based on annotated corpora and high-resource languages. However, some models have used the models related to rich languages and generalized them to low-resource ones using zero-short, few-shot or transfer learning.

In less-resource environments, which lack suitable annotated data, sentiment analysis can be performed using methods such as transfer learning [14, 18, 29], semi-supervised learning [25, 31, 32], and unsupervised learning [23, 47]. In these models, techniques such as semi-supervised manifold regularization [24, 25], methods based on recursive automatic encoders [39, 55], and hidden latent variable models can be utilized. Also, a group of researchers has used machine translation systems to translate other languages into English, and to use English sources in sentiment analysis of low-resource languages [22, 49, 50]. Such techniques facilitate sentiment analysis in multilingual and cross-lingual setups [40].

Nevertheless, evaluating sentiment analysis systems requires annotated data by native speakers due to various cultural and linguistic observations that should be taken into account for each language. With these explanations, it seems that for a low-resource language like Kurdish, producing an annotated corpus and creating an opinion analysis model based on it is necessary and needed.

In this paper, in contrast to these previous works, we first introduce an annotated corpus and then use several simple methods to evaluate it. In the same spirit of works applying transfer learning, we also rely on a pretrained model to leverage information from a richly-resourced language, i.e. English, for Kurdish sentiment analysis in this work.

## 3 METHODOLOGY

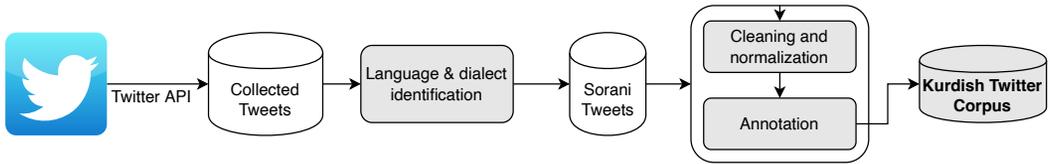

Fig. 1. A schema of data collection and annotation for Central Kurdish sentiment analysis

### 3.1 Data Collection

We selected Twitter as our source of data. Using Twitter API[2], we collected a few thousand tweets and filtered out the potentially Kurdish ones using regular expressions based on distinctive characters such as ڕ and ڵ along with fastText language identifier [35]. Furthermore, only tweets that are correct with regard to grammar or spelling are included. In cases where a slight modification in the tweet structure (but not wording) can make it conform to this condition, the annotator is asked to apply modifications as long as rewriting the whole tweet is not required. In addition, poems, famous quotations, and code-switched tweets are removed. The tweets are manually verified that the tweets are written in the Perso-Arabic script of Central Kurdish. Finally, we use KLPT [3] for text preprocessing and orthography normalization.

### 3.2 Data Annotation

The annotation process was carried out by two annotators native of Central Kurdish. The annotation was carried out based on an annotation guide where the selection of appropriate tweets and how to annotate them are described. The annotation is aimed at document-level sentiment classification to find the general sentiment of the author in an opinionated text, including emojis. Therefore, annotators determine a tweet as "subjective" or "objective" first, then, in the case of subjectivity, a label among the following ones is selected: positive, negative, mixed, neutral, and none. Usually, a subjective sentence is supposed to represent sentiment, not an objective one. Overall, 1769 instances are annotated.

In order to evaluate the quality of the annotations, we calculate the inter-annotator agreement (IAA) with Krippendorff's alpha [28] for nominal sentiment labels between the two annotators. As indicated in Table 1, the annotations achieve 0.84 of IAA in sentiment analysis. In other words, annotators agree on 84% of the labels they were expected to disagree on by chance. Krippendorff's alpha provides a useful measure of how often labels from different annotators agree in such a way that isolates annotators' skills. Finally, the annotations are aggregated by a third annotator where common annotations are finalized and those that are in conflict, are rectified. For instance, if one annotator marks a tweet as subjective and the other one as objective, it is considered mixed.

---
[2]Used in June 2021

It is worth mentioning that the annotation campaign initially intended to include hate speech detection where tweets were tagged if offensive and targeting individuals or groups. This, however, was a more challenging task with low agreement among annotators and a considerable imbalance and bias. Therefore, we did not include hate speech detection in this project. Detection of hate speech is particularly challenging as hatefulness is oftentimes a relative notion. For instance, the remarks of an atheist may be intellectual and factual but be considered hate speech toward religious groups. Similarly, how religious practitioners describe social issues and natural phenomena might be aggressive and offensive utterances.

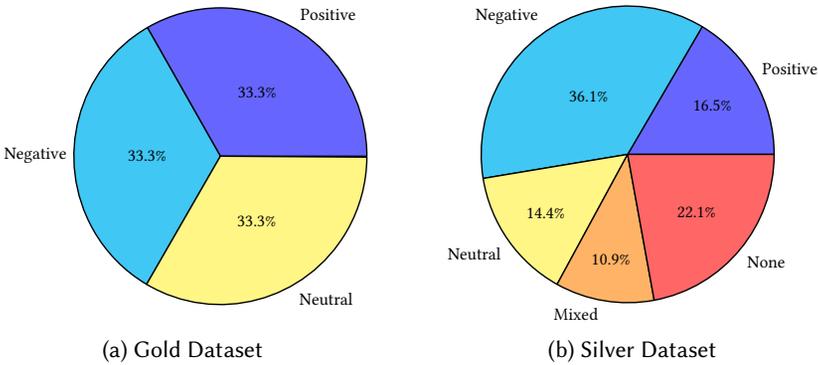

Fig. 2. Distribution of the instances among the classes in our gold-standard and silver-standard datasets used for sentiment analysis of Central Kurdish.

## 3.3 Transfer Learning

Robust classification systems require a considerable set of data. Given the small size of the annotated data, henceforth referred to as gold-standard dataset, due to limitations in time and cost, we also extend the data for Central Kurdish sentiment analysis using transfer learning [54]. Transfer learning refers to a set of methods where a model developed for a task is reused as another task; it has been previously used in low-resource NLP [56], particularly in sentiment analysis [15]. In our approach, we rely on a neural machine translation model pretrained on many languages including Central Kurdish and English along with a sentiment analysis model pretrained on English.

To that end, we use the No Language Left Behind (NLLB) [20] to translate the collected tweets to English described in §3.1, excluding those that are annotated and included in the gold-standard dataset. Following this, we determine the translated tweets with their sentiment labels using a RoBERTa-base model trained on 124M tweets in English from January 2018 to December 2021 [34].[3] Additionally, we made sure that out of the 1500 tweets, at least 500 tweets in each sentiment class (positive, negative, and neutral) contained an emoji. This was done to allow us to examine the impact of emojis on the performance of our models.

We believe that the resulting dataset, henceforth called the silver-standard dataset, can remedy the paucity of data for Central Kurdish sentiment analysis. The number of instances per class is provided in Table 2.

---

[3]Latest version of 2022 available at https://huggingface.co/cardiffnlp/twitter-roberta-base-sentiment-latest.

| Dataset | Positive | Negative | Neutral | Total |
|---|---|---|---|---|
| Gold-standard | 292 | 639 | 254 | 1185 |
| Silver-standard | 1500 | 1500 | 1500 | 4500 |

Table 2. Number of instances in the annotated dataset (gold-standard) and the silver-standard one created using transfer learning

## 3.4 Classification Algorithms

We train a total of five mainstream classification algorithms that are commonly used in sentiment analysis tasks due to their effectiveness. Four of these algorithms, namely logistic regression, decision trees, random forest tree, and support vector machines, are classical machine learning methods. On the other hand, a deep learning method is employed using a bi-directional long short-term memory model. Datasets are split into train and test with an 80%-20% ratio.

*3.4.1 Logistic Regression (LR).* This is a statistical method that we used to classify data into one of two categories [43]. It is a linear model that is simple to implement and can be used for the multiclass classification task. In this study, we used LR to classify tweets into positive, negative, or neutral classes. To optimize the performance of the model, we set the hyperparameters, including the optimizer to L-BFGS and the multinomial loss function, i.e. softmax.

*3.4.2 Decision Trees (DT).* This is a popular algorithm used for classification tasks [38]. It works by recursively partitioning the data into subsets based on the values of the input features.

*3.4.3 Random Forest (RF).* This is an ensemble method that combines multiple decision trees to improve the performance of the model [17]. By averaging the predictions of multiple DTs, RF reduces the overfitting and variance of a single DT. We utilize cross-entropy as the loss function and set the number of estimators to 30.

*3.4.4 Support Vector Machines (SVM).* This works by finding the hyperplane that maximally separates the data into different classes [41]. We used Linear SVC, a type of SVM, which operates by identifying the hyperplane that can best divide the data points into distinct categories. This approach is based on maximizing the margin between the data points and the hyperplane, ensuring that the model can accurately classify future observations.

*3.4.5 Bidirectional Long Short-Term Memory (BiLSTM).* This is a type of recurrent neural network (RNN) that is able to process sequential data by considering past and future contexts [26]. This makes it a powerful tool for our task. We employ embeddings to represent the data, wherein the embedding size is set to 100. The network architecture consists of two bidirectional layers, with sizes 64 and 32, respectively. We also include a dropout of 0.3 between the layers to prevent overfitting. Finally, we apply the softmax activation function to the last dense layer.

To train each of the classical machine learning models, we used both the text and non-text features of the tweets as inputs. These features were represented using the term frequency-inverse document frequency (TF-IDF) [51] representation method.

It is worth mentioning that due to the high imbalance among the five labels, we only include 'positive', 'negative', and 'neutral' in the classification task. This limitation can be further studied in future work. The distribution of different classes is illustrated in Figure 2. A few examples are provided in Table A.1 in Appendix.

## 4 EVALUATION

### 4.1 Evaluation Metrics

We use several common evaluation metrics to evaluate the performance of our models, namely accuracy, $F_1$ score, precision, and recall as described by Powers [45] as follows:

- Accuracy is the ratio of the number of correct predictions to the total number of predictions.
- Precision is the ratio of true positive predictions to the total number of positive predictions.
- Recall is the ratio of true positive predictions to the total number of actual positive instances.
- $F_1$ score is the harmonic mean of precision and recall.

### 4.2 Results

*4.2.1 Baseline system.* As a baseline system, we evaluate the performance of the models by testing on the gold-standard dataset without including emojis and training on the baseline, upsampled and merged training sets. The experiment results are presented in Table 3. The baseline results indicate that SVM achieves the highest accuracy of 56% and $F_1$ score of 53% among all models. This said the BiLSTM model outperforms all of the classical models in accuracy, achieving an accuracy of 57%. On the other hand, the analysis of the $F_1$ score, precision, and recall metrics reveals a remarkable imbalance among all models indicating a bias toward the negative class, which has the highest number of samples (639). This highlights the need to address data imbalance and improve the model's ability to distinguish between the different sentiment classes.

Although data augmentation in the unsample and merged systems does not substantially improve the $F_1$ score and accuracy in general, the transfer learning approach (twitter-robert model and NLLB system in Table 3) increases the baseline with a 0.54 $F_1$ score and 0.53 accuracy.

*4.2.2 Upsample system.* As the second system, we proceed to upsample the gold-standard dataset by incorporating additional samples from the silver-standard dataset. As such, we increase the number of samples to 700 instances per class to reduce the imbalance between classes. This leads to an improvement in performance for all the models, as demonstrated in Table 3. LR achieves an accuracy of 59%, which is a notable improvement. Surprisingly, the BiLSTM model experiences a decline in performance, although the $F_1$ score, recall, and precision metrics are more balanced and exhibit similar values. This implies a superior ability to generalize to all classes, in contrast to the previous evaluation that was conducted on the imbalanced data.

*4.2.3 Merged system.* In a last experimental setting, we merge both the gold and silver standard datasets, while still maintaining a balanced distribution of 1700 samples per class. We observe an overall improvement in performance, with more balanced metrics indicating better generalization across all classes. SVM achieves the highest accuracy and $F_1$ score of 61%. The performance of all the systems in the three setups of baseline, upsample and merged, respectively denoted by the dataset sizes of 1185, 2100 and 5100 instances, is shown in Figure 3. The $F_1$ score improves gradually by the size of datasets in all the systems, except DT. There could be various factors contributing to this issue, such as the quality of the combined dataset which may include noise, and the need to optimize the hyperparameters to ensure effective generalization for the larger dataset.

To conclude, the upsampling technique using the silver-standard dataset results in a considerable improvement in performance by at least 8%, while also achieving a better balance between $F_1$ score, recall, and precision, indicating better generalization. One notable observation is that the performance of a BiLSTM model may be highly dependent on the specific task and dataset, and may not always outperform classical machine learning models.

| Model | Test set | System | Precision | Recall | $F_1$ | Accuracy |
|---|---|---|---|---|---|---|
| LR | Baseline | Baseline | 0.49 | 0.54 | 0.44 | 0.54 |
|  |  | Upsample | 0.45 | 0.42 | 0.43 | 0.42 |
|  |  | Merged | 0.48 | 0.48 | 0.48 | 0.48 |
|  | Upsample | Upsample | **0.61** | 0.59 | 0.59 | 0.59 |
|  | Merged | Merged | **0.61** | 0.6 | 0.6 | 0.6 |
| SVM | Baseline | Baseline | 0.53 | 0.56 | 0.53 | 0.56 |
|  |  | Upsample | 0.47 | 0.45 | 0.45 | 0.45 |
|  |  | Merged | 0.47 | 0.47 | 0.47 | 0.47 |
|  | Upsample | Upsample | 0.54 | 0.53 | 0.53 | 0.53 |
|  | Merged | Merged | **0.61** | **0.61** | **0.61** | **0.61** |
| RF | Baseline | Baseline | 0.48 | 0.54 | 0.46 | 0.54 |
|  |  | Upsample | 0.52 | 0.47 | 0.48 | 0.47 |
|  |  | Merged | 0.47 | 0.44 | 0.45 | 0.44 |
|  | Upsample | Upsample | 0.55 | 0.52 | 0.52 | 0.52 |
|  | Merged | Merged | 0.55 | 0.55 | 0.55 | 0.55 |
| DT | Baseline | Baseline | 0.45 | 0.44 | 0.44 | 0.44 |
|  |  | Upsample | 0.46 | 0.42 | 0.43 | 0.42 |
|  |  | Merged | 0.42 | 0.37 | 0.39 | 0.37 |
|  | Upsample | Upsample | 0.5 | 0.5 | 0.5 | 0.5 |
|  | Merged | Merged | 0.48 | 0.48 | 0.48 | 0.48 |
| BiLSTM | Baseline | Baseline | 0.56 | 0.26 | 0.36 | 0.57 |
|  |  | Upsample | 0.44 | 0.41 | 0.44 | 0.46 |
|  |  | Merged | 0.42 | 0.42 | 0.44 | 0.44 |
|  | Upsample | Upsample | 0.59 | 0.52 | 0.55 | 0.53 |
|  | Merged | Merged | 0.56 | 0.57 | 0.56 | 0.57 |
| twitter-roberta | Baseline | NLLB | 0.53 | 0.53 | 0.54 | 0.53 |

Table 3. Performance of our systems with different test sets and setups without including emojis. Data augmentation improves the results (the highest scores are specified in bold).

## 4.3 Ablation Analysis

As an ablation analysis, we evaluate the impact of emojis on sentiment analysis for Central Kurdish. To that end, we train and test under the same setups presented in Table 3 with emojis in the tweets.

The experiment results presented in Table 4 indicate that the models achieve superior results when analyzing only the text of the tweets, without including emojis. We observe that the inclusion of emojis decreases the performance of all models by 2-3%. This could be attributed to potential inconsistencies between the sentiment conveyed by the emojis and the actual sentiment expressed in the tweet. Similarly, the presence of emojis continues to decrease the performance of all classical ML models except the BiLSTM, which is a neural network-based model. Neural networks tend to be more data-hungry and may benefit from the increased amount of data resulting from the dataset merge.

Nevertheless, the impact of emojis on performance is only evident when the number of samples is increased. Increasing the size of the dataset leads to more resilient models to emojis. While emojis tend to decrease the performance of classical ML models, they actually improve the performance of only the BiLSTM model.

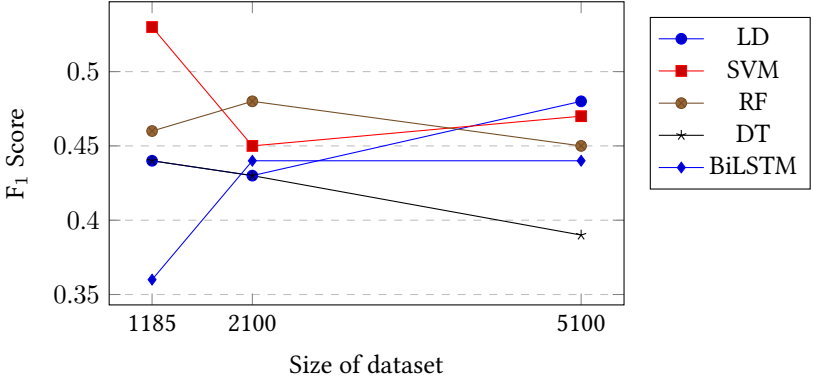

Fig. 3. Performance of various systems (without emojis) in different setups that are tested on the gold-standard data (20% of the 1185 instances) and trained on the gold-standard train set (1185 instances), up-sampled data (2100 instances) and those two merged (5100 instances), all with an 80-20 train-test ratio.

| Model | System | Precision | Recall | $F_1$ | Accuracy |
|---|---|---|---|---|---|
| LR | Baseline | 0.49 | 0.54 | 0.43 | 0.54 |
|  | Upsample | 0.59 | 0.58 | 0.58 | 0.58 |
|  | Merged | **0.62** | **0.61** | **0.61** | **0.61** |
| SVM | Baseline | 0.53 | 0.55 | 0.53 | 0.55 |
|  | Upsample | 0.53 | 0.53 | 0.53 | 0.53 |
|  | Merged | 0.58 | 0.57 | 0.57 | 0.57 |
| RF | Baseline | 0.46 | 0.54 | 0.42 | 0.54 |
|  | Upsample | 0.51 | 0.5 | 0.5 | 0.5 |
|  | Merged | 0.57 | 0.56 | 0.57 | 0.56 |
| DT | Baseline | 0.46 | 0.45 | 0.46 | 0.45 |
|  | Upsample | 0.46 | 0.45 | 0.45 | 0.45 |
|  | Merged | 0.48 | 0.48 | 0.48 | 0.48 |
| BiLSTM | Baseline | 0.51 | 0.5 | 0.5 | 0.51 |
|  | Upsample | 0.49 | 0.48 | 0.49 | 0.49 |
|  | Merged | 0.59 | 0.57 | 0.58 | 0.58 |

Table 4. Performance of our systems in different setups with data containing emojis.

## 5 CONCLUSION AND FUTURE WORK

In this study, we address the task of sentiment analysis for Central Kurdish with the primary goal of providing a benchmark. We describe the collection and annotation of data based on tweets and also, another data augmentation approach using transfer learning to automatically generate sentiment analysis instances. This dataset is openly available and can pave the way for future developments for Central Kurdish. To train and evaluate our models, we employ a variety of classical machine learning techniques, namely logistic regression, decision trees, random forest tree, and support vector machines along with the neural network-based model of bi-directional long short-term memory.

Comparing the performance of different techniques under various setups, we demonstrate that data augmentation is beneficial to increase the accuracy and $F_1$ score remarkably. We also carry out an ablation analysis and find that sentiment analysis without the actual emoji characters can slightly improve the results.

One of the limitations of the current study is the number of sentiment labels being positive, negative and neutral. This along with the lack of further annotated resources, particularly a polarity lexicon or translation of emojis, can be addressed in the future by incorporating lexicon-based information, as suggested by Lemmens et al. [30]. Exploring cross-lingual approaches for sentiment analysis of other Kurdish varieties and dialects is also suggested as a future work.

## ACKNOWLEDGMENTS

The authors would like to thank Behshad Davoudi, Iman Ghavami, and Arvin Rasooli at the University of Kurdistan for annotating the sentiment analysis data in Central Kurdish.

# A APPENDIX

We showcase a few examples of sentiment analysis predictions by the BiLSTM model. Specifically, we present examples of both correct and incorrect predictions made by the model, in order to provide insight into its strengths and weaknesses. By examining these examples, readers can gain a deeper understanding of the challenges and opportunities involved in sentiment analysis for Central Kurdish, and the potential limitations of current machine learning models in this domain.

| Reference | Prediction | Tweet |
| --- | --- | --- |
| Negative | Negative | ئاخ منالّی سپۆیلد و هیچ نەدیو چەن تێنەگەشتوو چەن ناشیرین چەن بێ سوود. <br> Oh, how ugly, useless and fool (is) a spoiled and bad-mannered kid. |
| Neutral | Negative | یادی بە خێر 😄 بە منالّی خەونەکانمان چەند گەورە بوون 😄😄💔💔💔 <br> Those were the days 😄 How big our childhood dreams were 😄😄💔💔💔 |
| Neutral | Positive | عەشقی ڕاستەقینە وەک و نوێژ وایە، دوای ئەوەی نیەتت هێنا نابێ سەیری دەور و بەرت بکەی. <br> Real love is like prayer. You should not get distracted when doing it. |
| Negative | Negative | بە دڕۆی پیاوە گەورەکان ئەڵێن سیاسەت <br> The big lies of big men (people) are called politics |
| Positive | Positive | ئەو کەسانەی بەقسەم ئەکەن لام زۆر شیرینننننن <br> I find those who listen to me so sweeeeeeet |
| Positive | Neutral | دڵّت بۆ لێدان دروست کراوە و ڕوخسارت بۆ پێکەنین دروست کراوە و ژیانیشت تەنیا موڵکی خۆتە. <br> Your heart is created to beat and your face to smile and your life is yours. |
| Positive | Negative | ئاشق بوون ئەرکی پیاوە، ژن خۆی عیشقە. 💛 <br> Falling in love is man's work. Woman is love. 💛 |
| Neutral | Neutral | تۆ سەرنجی ئەم هەموو هاتن و چوونی ئینسانانە بدەن؛ وەکوو خۆرمان لێ هاتووە، ئەم ئاوا ئەبێت و ئەوی تر هەڵدێت. <br> Look how humans arrive and leave. We are like the sun: coming and going. |

Table A.1. A few examples in our annotated dataset with references and BiLSTM model predictions. Code-switched words like سپۆیلد 'spoiled' and incorrect spellings are highlighted in red and yellow respectively. The translations are free. Note that these sentences are selected as examples and do not reflect authors' opinions.